\documentclass{article}

\usepackage{xcolor}
\usepackage[final]{corl_2026}
\usepackage{amsmath,amsfonts,amssymb,booktabs,array,graphicx,enumitem,float,wrapfig}
\usepackage{multirow}
\usepackage{makecell}
\usepackage{adjustbox}
\usepackage{microtype}
\usepackage{url}
\usepackage{caption}

\makeatletter
\renewcommand{\@conferencelocation}{Austin, Texas, USA}
\makeatother

\hypersetup{
  pdftitle={Robots that Collaborate: Sequential Asymmetric Imitation for Learning Coupled Robot Policies},
  pdfauthor={Ranpeng Qiu, Yincong Chen, Zihao Li, Yanan Zhou, Guoqiang Ren, Weiming Zhi}
}

\title{Robots that Collaborate: Sequential Asymmetric Imitation for Learning Coupled Robot Policies}

\author{
  Ranpeng Qiu\textsuperscript{1} \And
  Yincong Chen\textsuperscript{1} \And
  Zihao Li\textsuperscript{1,3} \And
  Yanan Zhou\textsuperscript{3} \And
  Guoqiang Ren\textsuperscript{1,2} \And
  Weiming Zhi\textsuperscript{3} 
}

\begin{document}
\maketitle
\begin{abstract}
Collaborative mobile manipulation requires robots to coordinate with a partially observed partner while physically interacting through shared objects. This is difficult because failures often arise not from poor local skills, but from mistimed waiting, yielding, pulling, releasing, or repositioning. We study this problem with two bimanual mobile manipulators coupled through rigid and deformable objects. We propose \textit{Sequential Asymmetric Imitation} (SAI), a staged imitation curriculum that can be collected through a single teleoperation interface, without synchronized dual-robot teleoperation or explicit inter-robot communication. SAI trains Robot A from unilateral demonstrations with a compliant human partner, trains Robot B against the deployed Robot A policy, and then refines Robot A using sparse interventions near coordination failures. This staged process exposes the policies to increasingly realistic partner behaviors, including delay, phase mismatch, insufficient yielding, and interaction conflict. Across real-world dual-robot manipulation tasks, SAI improves task success, phase synchronization, and partner-contingent yielding over independent imitation and curriculum-ablation baselines. These results suggest that physically coupled collaboration can be learned through the structure of the imitation curriculum, rather than through synchronized multi-operator demonstrations or explicit coordination mechanisms.
\end{abstract}
\keywords{Mobile Manipulation, Imitation Learning, Policy Learning}
\footnotetext[1]{Zeno AI. }\footnotetext[2]{Zhejiang University. }\footnotetext[3]{The University of Sydney.}


\section{Introduction}

Dual-robot manipulation enables robots to handle large, heavy, and deformable objects that are difficult for a single robot to control. However, success in this setting requires more than two locally competent policies. When robots pull, lift, align, transport, or spread a shared object, each robot's motion changes the load, contact state, and feasible motion of the other. Small timing errors can therefore produce large interaction forces, object deformation, loss of grasp, or task failure. Recent work has made substantial progress in robot manipulation from demonstrations, including mobile manipulation, dual-arm manipulation, and visuomotor policy learning with expressive architectures such as action chunking transformers and diffusion policies~\cite{zhao2023learning,chi2023diffusion,fu2024mobile}. Yet physically coupled dual-robot tasks introduce a distinct coordination problem. The robots must align task phases, adapt to partner delays, yield under interaction conflict, and resume coordinated motion after perturbations. Independent policies often fail because each robot follows its own nominal motion phase, even when the partner is delayed, stalled, or out of position.

\begin{figure}[!t]
\centering
\includegraphics[width=\linewidth]{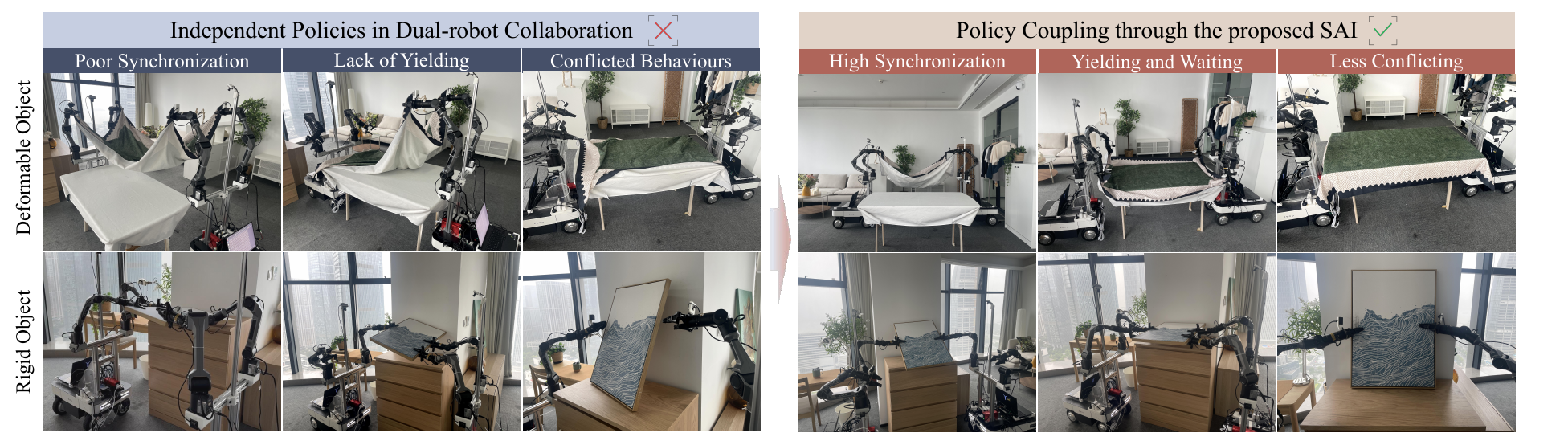}
\caption{
\textbf{Local skills are not enough for dual-robot coordination.}
Independent policies can fail under phase mismatch, insufficient yielding, and interaction conflict. SAI induces coupled behavior.
}
\label{fig:teaser}
\end{figure}

Existing approaches face practical bottlenecks in this setting. Model-based planning and centralized control can coordinate multiple robots when object geometry, contacts, and interaction constraints are well specified~\cite{devincenti2023centralized, GeoFab_gloabL_opt,liu2025planning,lai2022parallelised}, but these assumptions are hard to maintain with deformable objects, intermittent contacts, friction, and long-horizon mobile manipulation. Multi-agent reinforcement learning provides a general coordination framework~\cite{lowe2017multi}, but real-robot deployment remains difficult due to sparse rewards~\cite{ravichandar2020recent}, reward shaping, and simulation-to-real transfer. Imitation learning is more direct, but standard dual-robot imitation typically requires synchronized joint demonstrations, where both robots are controlled as a coordinated pair. Collecting such data requires multiple teleoperation channels, calibrated interfaces, and operators who can maintain precise temporal alignment~\cite{zhi2024sketch}. This data-collection requirement motivates our central question: \emph{Can coupled dual-robot policies be learned from single-teleoperator demonstrations, without synchronized dual-operator data, reinforcement learning, or explicit inter-robot communication?}

We propose \textit{Sequential Asymmetric Imitation} (SAI), a staged imitation curriculum for learning physically coupled dual-robot behavior. SAI decomposes coordination learning into three asymmetric stages. First, Robot A learns basic task execution from unilateral demonstrations with a compliant human or passive partner. Second, Robot A is deployed as a frozen policy while a single operator teleoperates Robot B, allowing Robot B to learn against the timing, motion profile, and errors of its future robot partner. Third, both policies are deployed together, and sparse interventions near coordination failures refine Robot A with behaviors such as waiting, yielding, slowing, and recovery.

Our key hypothesis is that inter-robot coordination can be induced by structuring the partner distribution seen during imitation learning. SAI shifts training from compliant support, to a deployed learned partner, to closed-loop robot-robot execution with targeted corrections. This exposes the policies to realistic coordination errors, including partner delay, phase mismatch, insufficient yielding, and interaction conflict, enabling coordination through local observations and shared physical interaction without explicit communication or centralized action prediction.

Concretely, our technical contributions are:
\begin{enumerate}[leftmargin=*,nosep]
\item \textbf{Sequential Asymmetric Imitation.} We introduce SAI, a single-teleoperator curriculum for learning physically coupled dual-robot policies without synchronized dual-operator demonstrations, reinforcement learning, or explicit inter-robot communication.

\item \textbf{Partner-distribution curriculum.} We show that shifting the training partner from compliant support, to a frozen learned policy, to closed-loop robot-robot execution induces phase alignment and partner-contingent yielding.

\item \textbf{Real-world dual-robot validation.} We evaluate SAI on contact-rich real-world tasks involving rigid and deformable shared objects, showing improvements in task success, phase synchronization, and yielding behavior over independent imitation and curriculum-ablation baselines.
\end{enumerate}

\section{Related Work}

\textbf{Dual-Robot and Multi-Arm Manipulation:}
Classical methods coordinate multiple manipulators by modeling object geometry, contacts, and kinematic constraints, but these assumptions weaken with deformable materials, intermittent contacts, and unmodeled interaction dynamics~\cite{aksoy2024collaborative,hannus2024dynamic}. Learning-based methods reduce manual modeling through visuomotor imitation~\cite{kim2024goal,zhi2022diffeomorphic, Fast_diff_int}, including diffusion policies, keypoint representations~\cite{gao2024bikvil}, and bimanual foundation models~\cite{liu2025rdtb}. Most, however, rely on synchronized joint trajectories that supervise both agents as one coordinated behavior~\cite{shaw2024bimanual,jiang2025rethinking}. This increases data-collection cost, entangles agent roles, and can make policies sensitive to partner timing deviations~\cite{lin2024twisting}. SAI instead induces dual-robot coordination through staged partner-distribution shifts using single-teleoperator, asynchronous demonstrations.
\vspace{-0.3em}

\textbf{Teleoperation-Based Imitation Learning:}
Teleoperation-based imitation methods such as ACT~\cite{zhao2023learning}, Diffusion Policy~\cite{chi2023diffusion}, and flow-based methods \cite{long2026constraining} achieve strong visuomotor manipulation performance, but collaborative dual-robot tasks require demonstrations that capture temporal coupling between agents. Existing multi-arm systems often use stationary bimanual setups or specialized capture and teleoperation interfaces~\cite{fu2024mobile,chi2024universal} to obtain aligned trajectories, increasing hardware and operator requirements. Recent work also studies efficient data collection and dataset scaling for robot manipulation~\cite{gao2024efficient,saxena2025what}, but synchronized collaborative data on mobile manipulators, studied in this work, remain expensive to collect. SAI assumes joint demonstrations are unavailable and bootstraps coordination through a sequential, asymmetric \emph{single}-teleoperator curriculum.
\vspace{-0.3em}

\textbf{Interactive Imitation and Intervention Learning:}
Robot learning from demonstration has studied how skills can be acquired from human examples through general imitation-learning formulations~\cite{osa2018algorithmic}, dynamical-system-based policies~\cite{ravichandar2017contracting, periodic}, and synchronized movement primitives for bimanual manipulation~\cite{thota2016synchronization}. These approaches typically assume that the relevant coordination structure is present in the demonstrations or specified through primitive-level synchronization. Interactive imitation methods, including DAgger and intervention-based variants, reduce covariate shift by collecting corrections under the learner-induced state distribution~\cite{ross2010reduction,hoque2023interventional}. SAI uses this principle only in its final stage, but targets coordination-specific failures such as partner delay, phase mismatch, and near-failure contact configurations~\cite{spencer2020learning}. Targeted loss masking and nominal replay preserve existing skills while adding partner-contingent yielding and recovery. Recent related approaches such as R2BC, SoCo, and MART also study
learning and refinement mechanisms relevant to coordinated robot behavior
~\cite{R2BC_KEY,SOCO_KEY,MART_KEY}. Our contribution is complementary:
SAI does not introduce a new online imitation-learning update, but instead
specifies a partner-distribution curriculum---which robot is supervised,
against which deployed partner, and at which stage---for physically coupled
dual-robot learning under a single teleoperation interface.
        
\section{Problem Formulation}

\textbf{Dual-Robot Imitation Learning Setup:} We formulate cooperative dual-robot manipulation as a decentralized partially observable control problem with global state $s_t \in \mathcal{S}$. At each timestep $t$, each robot $i \in {A,B}$ receives a local visuo-proprioceptive observation $o_i^t = O_i(s_t)$, and acts according to its own policy:
\begin{equation}
a_A^t \sim \pi_A(\cdot \mid o_A^t), \qquad
a_B^t \sim \pi_B(\cdot \mid o_B^t).
\end{equation}
The policies do not exchange messages, partner states, future actions, or latent embeddings. Coordination can only emerge through local observations of the shared workspace and the physical effects of manipulating the same object. Given demonstration data, the objective is to learn decentralized policies $\pi_A$ and $\pi_B$ by supervised imitation such that their joint rollout $\tau = {(o_A^t,o_B^t,a_A^t,a_B^t)}_{t=1}^{T}$ satisfies the task-success criteria with high probability.
\vspace{-0.3em}

\textbf{Collaboration Challenges:} Local skill does not imply cooperative skill. Even if each robot can grasp, lift, pull, or transfer objects in isolation, joint deployment can fail because both policies interact through the shared object and workspace. We focus on three recurring bottlenecks in physically coupled dual-robot manipulation:
\begin{itemize}[leftmargin=*,nosep]
\item \textbf{Temporal phase coupling:} the robots must align key task transitions, such as grasping, motion onset, lifting, lowering, and release, despite variable execution timing.
\item \textbf{Partner-contingent yielding:} each robot must slow, pause, or re-time its motion when the partner is delayed, stalled, obstructed, or out of phase.
\item \textbf{Interaction conflict mitigation:} the joint behavior must avoid excessive internal forces, object deformation, opposing motions, and workspace collisions.
\end{itemize}
We call this ability \textit{implicit policy coupling}: partner-responsive coordination that arises from local observations and shared physical interaction, without explicit inter-robot communication. A coupled policy pair should maintain phase-aligned progress, yield to partner deviations, and avoid interaction-induced failures during closed-loop execution.
\vspace{-0.3em}

\textbf{Single-Teleoperator Constraints:} Standard dual-robot imitation learning assumes synchronized joint demonstrations, where both robots are controlled and recorded as one coordinated trajectory. This can require multiple teleoperation channels, calibrated interfaces, and precise temporal alignment between operators. We instead impose a single-teleoperator constraint: only one robot can be directly teleoperated at a time through one hardware interface. Under this constraint, supervision is decomposed into three asymmetric datasets: $\mathcal{D}_A^{\mathrm{solo}}$teleoperate Robot A with a compliant human or passive partner; $\mathcal{D}_B^{A}$teleoperate Robot B while the learned Robot-A policy is deployed; $\mathcal{D}_A^{\mathrm{int}}$ intervene on Robot A during closed-loop dual-robot deployment. The superscript in $\mathcal{D}_B^{A}$ indicates that Robot B demonstrations are collected under the partner distribution induced by the learned Robot-A policy. The central question is whether these sequential, asymmetric datasets are sufficient to learn coupled decentralized policies that approach the cooperative behavior normally obtained from synchronized joint demonstrations.

\begin{figure}[t]
    \centering
    \includegraphics[width=\linewidth]{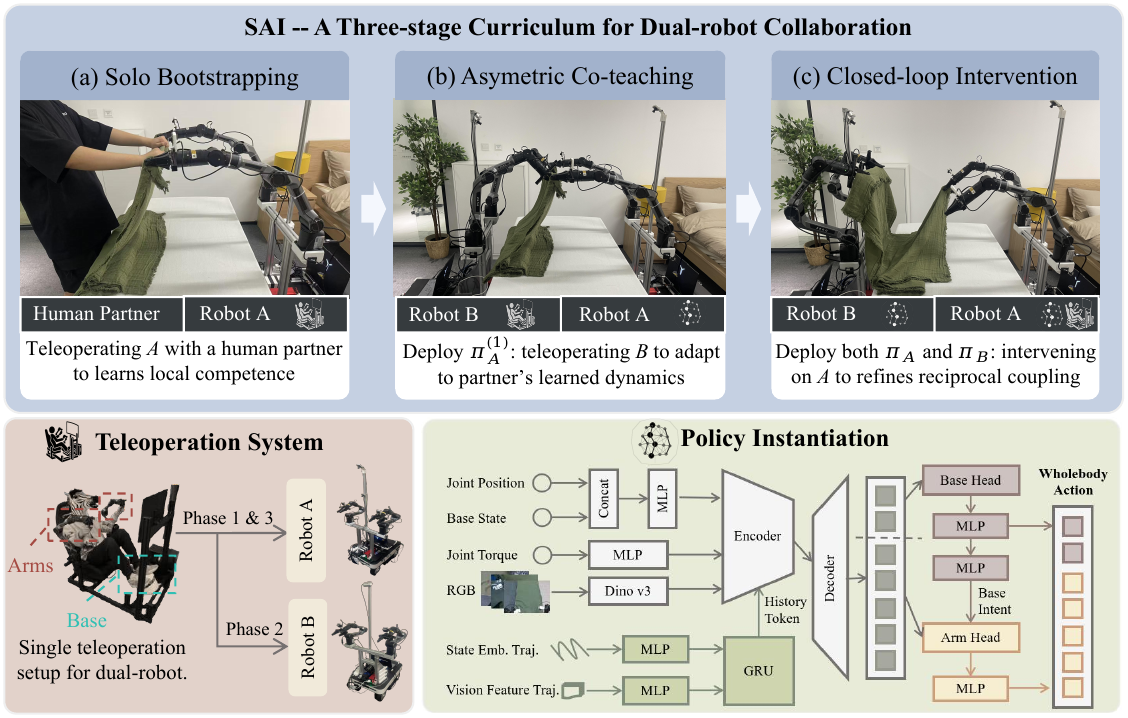}
    \caption{
    \textbf{System overview of SAI.}
    A single whole-body teleoperator~\cite{li2026tripilotff} collects a staged curriculum for policy coupling; bottom panels show the teleoperation interface and policy instantiation.
    }
    \label{fig:sai_overview}
\end{figure}
        
\section{Sequential Asymmetric Imitation}
\label{sec:sai}

Sequential Asymmetric Imitation (SAI) learns physically coupled dual-robot manipulation without requiring synchronized dual-operator demonstrations. The key idea is to build coordination gradually: first train one robot to perform the task with external support, then train the second robot against the deployed first robot, and finally correct the remaining closed-loop coordination failures through sparse human interventions.

SAI follows the data-collection sequence
\begin{align}
    \mathcal{D}_A^{\mathrm{solo}}
    \rightarrow
    \mathcal{D}_{B \mid A}^{\mathrm{demo}}
    \rightarrow
    \mathcal{D}_A^{\mathrm{int}} .
    \label{eq:sai_data_sequence}
\end{align}
Here, $\mathcal{D}_A^{\mathrm{solo}}$ contains unilateral demonstrations for Robot A, $\mathcal{D}_{B \mid A}^{\mathrm{demo}}$ contains Robot-B demonstrations collected while Robot A executes its learned policy, and $\mathcal{D}_A^{\mathrm{int}}$ contains corrective Robot-A interventions collected during joint deployment. Each robot acts from its own local history, including recent images, proprioception, and previous actions. The policies do not receive privileged partner state, explicit communication, or centralized coordination signals.

\subsection{Stage 1: Bootstrapping Robot A}
\label{sec:sai_stage1}

SAI begins with unilateral demonstrations for Robot A. In these demonstrations, Robot A performs the task while the other side of the shared object is supported by a compliant human or passive partner. This gives Robot A basic task competence, such as grasping, pulling, aligning, and exposing task-relevant regions, without requiring two robots or two synchronized operators.

Because the Stage-1 partner is human but the deployment partner is Robot B, a policy trained directly on these demonstrations may overfit to human-specific appearance and occlusion patterns. We therefore apply partner-masking augmentation to frames in which the supporting partner is visible: the partner region is replaced with randomized appearance perturbations, such as color jitter, blur, noise, or texture changes, with details in Appendix~\ref{app:partner_masking}. This encourages Robot A to focus on task-relevant object state rather than the visual identity of the supporting partner. Robot A is trained by behavioral cloning on the augmented demonstrations, producing the initial policy $\pi_A^{(1)}$. This policy can perform the task with external support, but it has not yet learned to coordinate with an autonomous robot partner.

\subsection{Stage 2: Training Robot B Against Deployed Robot A}
\label{sec:sai_stage2}

The second stage freezes $\pi_A^{(1)}$ and deploys it on Robot A. A single operator then teleoperates Robot B while interacting with the deployed Robot-A policy. This produces $\mathcal{D}_{B \mid A}^{\mathrm{demo}}$, where Robot B observes the partner behavior it will face at test time.

During data collection, Robot A is executed with bounded deployment variation, such as speed scaling, action noise, timing offsets, or perturbed initial object states. These variations expose Robot B to realistic partner deviations and encourage waiting, yielding, recovery, and resumption behaviors. Robot B is then trained by behavioral cloning on the teleoperated actions, producing $\pi_B^{(2)}$. This stage is asymmetric: Robot B is not trained against an ideal partner or a manually synchronized human partner, but against the actual learned behavior of Robot A. As a result, Robot B learns to compensate for Robot A's timing, motion profile, and residual errors.

\subsection{Stage 3: DAgger-Style Intervention Fine-Tuning}
\label{sec:sai_stage3}

After Stage 2, Robot B has learned to coordinate with Robot A, but Robot A has still only been trained with a human-supported partner. The largest remaining distribution mismatch is therefore on Robot A: it must adapt from a compliant human partner to the autonomous behavior of Robot B. In practice, the remaining failures often occur at coordination boundaries, where Robot A pulls too early, fails to yield, continues while Robot B is delayed, or recovers poorly after phase mismatch.

To correct these failures, we deploy $\pi_A^{(1)}$ and $\pi_B^{(2)}$ together and collect sparse human interventions on Robot A. The intervention process is DAgger-style: the learned policies induce the closed-loop states, and the operator provides corrective labels only near states where coordination begins to fail. Each intervention records the current Robot-A history and the corrective action supplied by the operator. The resulting intervention dataset $\mathcal{D}_A^{\mathrm{int}}$ is aggregated with the original Robot-A demonstrations to fine-tune Robot A:
\begin{align}
    \pi_A^{(3)}
    =
    \operatorname{BC}
    \left(
        \mathcal{D}_A^{\mathrm{solo}}
        \cup
        \mathcal{D}_A^{\mathrm{int}}
    \right).
    \label{eq:sai_stage3_bc}
\end{align}
In practice, intervention samples are upweighted relative to nominal replay so that Robot A preserves its original task skill while adapting near coordination-critical states. When only part of the action requires correction, we optionally apply an action-dimension mask so that the loss supervises only the corrected components; for example, an operator may correct the base motion while leaving the arm command unchanged. If no mask is used, the full corrective action is used as the supervision target. This update is A-centric because Robot B has already been trained against deployed Robot-A behavior, whereas Robot A has not yet been trained against an autonomous Robot-B partner. Stage 3 recalibrates Robot A to the coupled deployment distribution, teaching it to slow down, wait, yield, or recover when the joint system deviates from the nominal phase. Safety overrides may be applied to Robot B during data collection, but they are not used as training labels. To separate the effect of the curriculum from additional data, we match the per-robot trajectory budget across training pipelines. Robot A uses 56 Stage-1 trajectories and 24 Stage-3 correction trajectories, for 80 trajectories in total, while Robot B uses 80 Stage-2 trajectories. The Independent Imitation and Serial Imitation baselines likewise use 80 trajectories per robot. Detailed data and intervention statistics are reported in Appendix~\ref{app:intervention_statistics}.
\subsection{Policy Instantiation}
\label{sec:sai_policy}

SAI is a training curriculum and can be instantiated with different decentralized visuomotor policies. In our experiments, each robot uses an ACT-style action-chunking policy with temporal history encoding and a cascaded whole-body action head. The policy receives local RGB observations from the shared top-view camera and egocentric wrist cameras, together with the robot's proprioceptive state and previous actions. Visual features are extracted with a frozen DINOv3-80M backbone, projected into image tokens, and fused with state features.

To provide temporal context for coordination decisions, we maintain a history window of $H=30$ steps and select 12 frames using logarithmic temporal sampling. Multi-view visual features and full-body state features are fused and encoded by a single-layer GRU, whose final hidden state forms the history token. The ACT decoder then predicts a whole-body action chunk with a cascaded head: a base-action MLP first predicts $a_{\mathrm{base}}\in\mathbb{R}^3$, corresponding to $(v_x,v_y,\omega)$, and an arm-action MLP then predicts $a_{\mathrm{arm}}\in\mathbb{R}^{14}$ conditioned on both the decoder query and the predicted base action. This architecture improves local execution reliability by encouraging base-arm consistency and preserving temporal context. However, the partner-responsive behavior comes from the SAI curriculum itself: each robot learns from progressively more realistic partner behavior, rather than from explicit communication, privileged partner state, or a centralized coordination loss.

\section{Experimental Results}
\label{sec:experiments}

Our evaluation addresses four questions:
\textbf{(Q1) Task performance:} Does SAI improve collaborative success across contact-rich real-world dual-robot manipulation tasks?
\textbf{(Q2) Partner delays:} Does SAI induce yielding when the partner is unexpectedly delayed?
\textbf{(Q3) Policy components:} Which lightweight architectural components improve execution reliability, and do they explain the cooperative gains?
\textbf{(Q4) Backbone compatibility:} Does the SAI data curriculum remain effective with different imitation-learning backbones?

Across comparisons, we keep the observation space, action space, and policy architecture fixed whenever possible, so changes in collaborative behavior can be attributed primarily to the training curriculum. Unless otherwise stated, policies use the ACT-style whole-body architecture described in Appendix~\ref{appendix:implementation}: frozen DINOv3 visual features, a temporal history encoder over RGB, proprioception, and torque feedback, and a cascaded action head that predicts mobile-base commands before arm-gripper commands. Each robot acts from local observations only, with no explicit inter-robot communication, privileged partner state, or centralized coordination loss. We report three percentage metrics. \textit{Success} measures task completion. \textit{Phase Sync.} measures whether both robots complete predefined task checkpoints within the task-specific timing tolerance and without a disruptive interaction. These checkpoints are fixed before evaluation and are listed in Appendix~\ref{app:evaluation_protocol}. \textit{Yield/Wait} measures whether the robot slows, pauses, or resumes appropriately when its partner is delayed or out of phase. Phase Sync. and Yield/Wait are manually scored from pre-specified task events and partner-response opportunities using fixed task-specific criteria. To assess scorer dependence, an independent rescoring of a randomized subset yielded 93.4\% agreement and Cohen's $\kappa=0.856$. Full event definitions are provided in Appendix~\ref{app:evaluation_protocol}.

\subsection{Q1: Does SAI improve collaboration across real-world tasks? Yes!}
\label{sec:exp_q1}

\begin{figure}[!t]
    \centering
    \includegraphics[width=\linewidth]{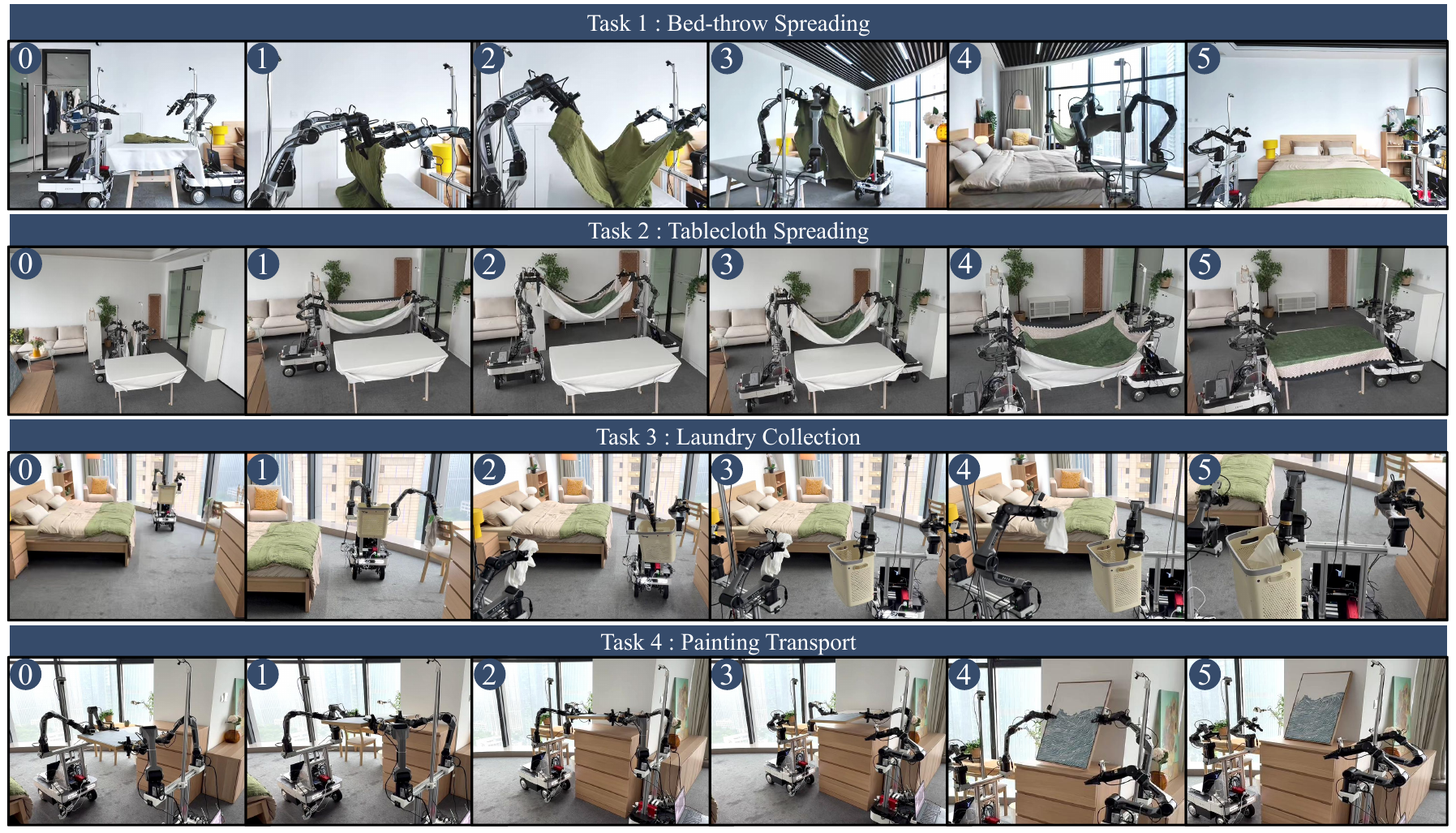}
    \caption{\textbf{Real-world task suite.} Representative rollouts across deformable spreading, shared-workspace collection, and rigid-object transport.}
    \label{fig:task_rollouts}
\vspace{-0.7em}
\end{figure}

We evaluate SAI on four real-world dual-robot manipulation tasks: \textbf{Bed-throw Spreading} and \textbf{Tablecloth Spreading}, which require deformable-object alignment; \textbf{Laundry Collection}, which requires asynchronous coordination in a shared workspace; and \textbf{Painting Transport}, which requires contact-rich rigid-object transport. As shown in Fig.~\ref{fig:task_rollouts}, these tasks cover phase mismatch, partner timing variation, shared-workspace interference, and unmodeled tracking latency.

\begin{wraptable}[15]{r}{0.58\linewidth}
  \centering
  \vspace{-0.55em}
  \captionsetup{font=small,skip=1.5pt}
  \caption{\textbf{Per-task evaluation.} SAI improves task success and process-level coupling metrics across deformable spreading, shared-workspace collection, and rigid-object transport.}
  \label{tab:per_task_results}
  \scriptsize
  \setlength{\tabcolsep}{2.2pt}
  \renewcommand{\arraystretch}{0.80}
  \begin{adjustbox}{width=\linewidth}
  \begin{tabular}{llccc}
  \toprule
  Task & Method & Success $\uparrow$ & Phase Sync. $\uparrow$ & Yield/Wait $\uparrow$ \\
  \midrule
  Bed-throw
  & Independent Imit. & 23.3\% & 30.8\% & 26.7\% \\
  & Serial Imitation & 36.7\% & 60.8\% & 35.0\% \\
  & \textbf{SAI} & \textbf{53.3\%} & \textbf{62.5\%} & \textbf{68.3\%} \\
  \midrule
  Tablecloth
  & Independent Imit. & 43.3\% & 54.4\% & 46.7\% \\
  & Serial Imitation & 60.0\% & 67.8\% & 61.7\% \\
  & \textbf{SAI} & \textbf{66.7\%} & \textbf{68.9\%} & \textbf{70.0\%} \\
  \midrule
  Laundry
  & Independent Imit. & 50.0\% & 51.7\% & 31.7\% \\
  & Serial Imitation & 63.3\% & 68.3\% & 48.3\% \\
  & \textbf{SAI} & \textbf{70.0\%} & \textbf{73.3\%} & \textbf{71.7\%} \\
  \midrule
  Painting
  & Independent Imit. & 33.3\% & 42.7\% & 34.4\% \\
  & Serial Imitation & 46.7\% & 67.3\% & 48.9\% \\
  & \textbf{SAI} & \textbf{56.7\%} & \textbf{74.7\%} & \textbf{68.9\%} \\
  \bottomrule
  \end{tabular}
  \end{adjustbox}
  \vspace{-0.45em}
\end{wraptable}

We compare three training pipelines under the same policy architecture. \textit{Independent Imitation} trains each robot from separately collected unilateral demonstrations. \textit{Serial Imitation} trains Robot B against the deployed Robot-A policy but omits the final intervention stage. Importantly, Serial Imitation
does not provide either policy with explicit partner state or
inter-robot communication. \textit{SAI} uses the full three-stage curriculum, including closed-loop Robot-A interventions.

Table~\ref{tab:per_task_results} shows that SAI improves task success and process-level coordination across all four tasks. Independent Imitation often learns local manipulation primitives but fails during joint execution because the robots follow incompatible nominal timings. Serial Imitation improves phase alignment by exposing Robot B to the deployed Robot-A policy; for example, Phase Sync. in Bed-throw Spreading increases from 30.8\% to 60.8\%. However, Robot A remains trained only under the human-supported Stage-1 distribution, limiting its ability to adapt when Robot B is delayed or out of phase. SAI further improves Success and Yield/Wait behavior, indicating that sparse closed-loop interventions teach Robot A partner-responsive correction behaviors without explicit communication or reinforcement learning.
\paragraph{Partner-informed controls.}
We additionally compare SAI against Diffusion Co-Policy (DCP) and a
Centralized Joint Diffusion Policy (CJDP), both of which use explicit
inter-robot information exchange. On Tablecloth and Painting, these
partner-informed policies are competitive under clean communication,
and SAI is not uniformly superior when perfect partner information is
available. However, their coordination degrades under a 100-ms
communication delay with 10\% packet loss, whereas SAI does not rely on
inter-robot communication. We therefore view SAI as a communication-free
alternative rather than as a universally superior replacement for
partner-informed control. Full results are reported in
Appendix~\ref{app:partner_controls}.

\subsection{Q2: Does SAI induce yielding under partner delay? Yes!}
\label{sec:exp_q2}

\begin{figure}[t]
\centering
\includegraphics[width=\linewidth]{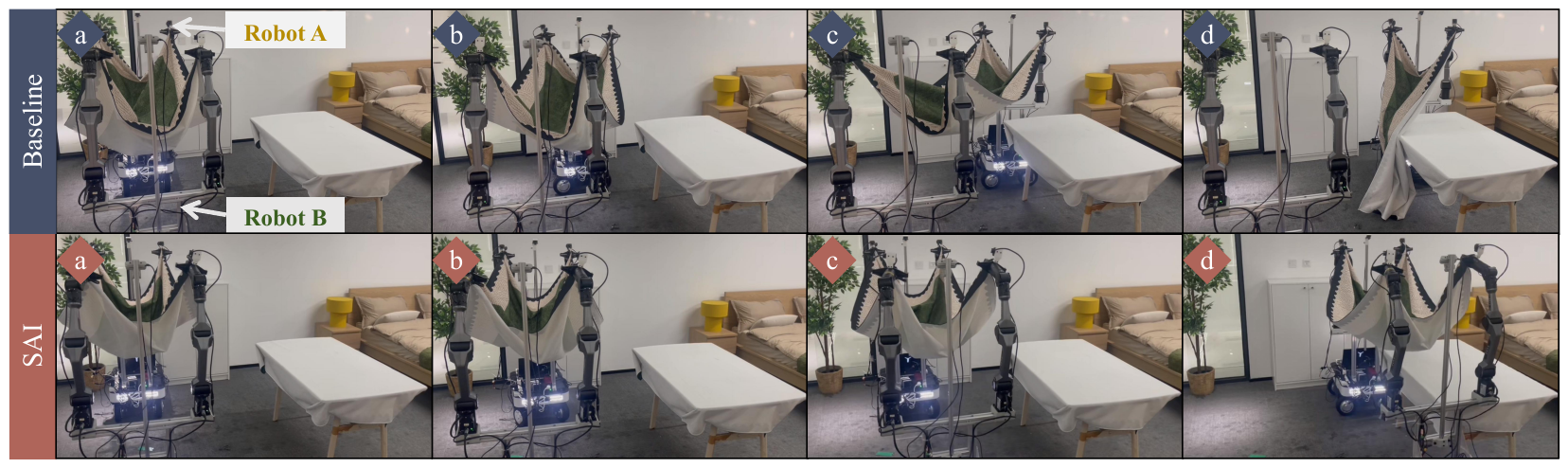}
\caption{
\textbf{Partner-contingent yielding under delay.}
When Robot B is paused in Tablecloth Spreading, Independent Imitation continues pulling, increasing cloth tension and causing Robot B to lose its grasp.
SAI instead slows, waits, and resumes after the partner recovers.
}
\label{fig:yield_response}
\vspace{-0.7em}
\end{figure}

To isolate partner-contingent yielding, we introduce a controlled delay during cooperative deployment in Tablecloth Spreading. Robot B is manually paused for a short interval, and we evaluate whether Robot A continues its nominal pulling motion or adapts to the delayed partner. This targets a common local coordination failure: if Robot A keeps pulling while Robot B remains stationary, the cloth becomes misaligned and the interaction can destabilize. Figure~\ref{fig:yield_response} shows representative rollout snapshots. The top row shows Independent Imitation, where Robot A continues its nominal motion despite Robot B being paused. This creates asymmetric pulling and leads to cloth misalignment. The bottom row shows SAI, where Robot A slows down and waits after observing that Robot B has not advanced, then resumes once the partner recovers. These snapshots illustrate the behavior captured by the aggregate Yield/Wait metric in Table~\ref{tab:per_task_results}: SAI increases partner-contingent slowing and waiting rather than simply replaying a nominal action sequence.

\subsection{Q3: Do history and cascading components in the policy improve reliability? Yes!}
\label{sec:exp_q3}

\begin{wraptable}[8]{r}{0.40\linewidth}
\centering
\vspace{-0.65em}
\captionsetup{font=small,skip=1.5pt}
\caption{Failure-specific policy ablations.}
\label{tab:policy_ablation}
\scriptsize
\setlength{\tabcolsep}{3pt}
\renewcommand{\arraystretch}{0.86}
\begin{adjustbox}{width=\linewidth}
\begin{tabular}{llc}
\toprule
Failure mode & Variant & Rate $\downarrow$ \\
\midrule
\multirow{2}{*}{Premature release}
& w/o Hist. & 64.5\% \\
& w/ Hist.  & \textbf{16.1\%} \\
\midrule
\multirow{2}{*}{Premature lowering}
& w/o Casc. & 51.6\% \\
& w/ Casc.  & \textbf{9.7\%} \\
\bottomrule
\end{tabular}
\end{adjustbox}
\vspace{-0.35em}
\end{wraptable}

SAI primarily changes the training distribution, but reliable real-robot deployment also depends on the policy's ability to execute whole-body actions consistently. We therefore ablate two components of the policy instantiation: the temporal history token and the cascaded base-arm action head. These ablations use the same sensorimotor interface and training pipeline, changing only the tested architectural component.

\begin{figure}[t]
\centering
\captionsetup{skip=2pt}
\begin{minipage}{0.5\linewidth}
\centering
\includegraphics[width=\linewidth]{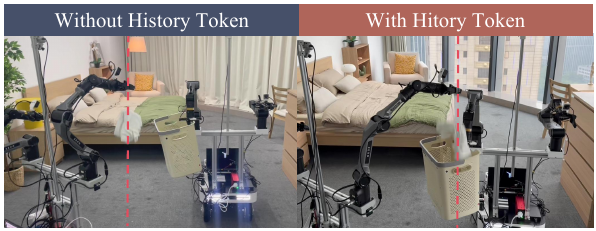}
\centerline{\small (a) Early release}
\end{minipage}
\hspace{-1em}
\begin{minipage}{0.495\linewidth}
\centering
\includegraphics[width=\linewidth]{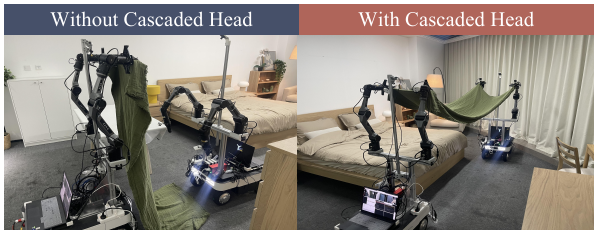}
\centerline{\small (b) Early lowering}
\end{minipage}
\caption{
\textbf{Policy ablations on execution failures.}
The history token reduces premature release, while the cascaded base-arm head reduces premature lowering by improving whole-body action consistency.
}
\label{fig:policy_ablation}
\vspace{-0.9em}
\end{figure}

Table~\ref{tab:policy_ablation} and Fig.~\ref{fig:policy_ablation} show that temporal history reduces premature release in Laundry Collection from 64.5\% to 16.1\%, because the policy can distinguish intermediate transport states from terminal release states. The cascaded base-arm head reduces premature lowering from 51.6\% to 9.7\%, because arm commands are conditioned on the predicted base motion rather than produced by an independent flat head. These components improve local execution reliability, but they do not explain the cooperative gains by themselves: the baseline pipelines in Table~\ref{tab:per_task_results} use the same architecture and still fail more often under partner delay and phase mismatch. Thus, the architecture stabilizes primitive execution, while SAI induces partner-responsive coordination through the data curriculum.

\subsection{Q4: Is SAI agnostic to specific imitation-learning backbones? Yes!}
\label{sec:exp_q4}
\begin{wraptable}[7]{r}{0.45\linewidth}
\centering
\vspace{-0.45em}
\captionsetup{font=small,skip=1pt}
\caption{\textbf{Backbone compatibility.} SAI improves over Independent Imitation.}
\label{tab:policy_backbone}
\scriptsize
\setlength{\tabcolsep}{1.6pt}
\renewcommand{\arraystretch}{0.82}
\begin{adjustbox}{width=\linewidth}
\begin{tabular}{llccc}
\toprule
Action Head
& Training Pipeline
& Success $\uparrow$
& Phase Sync. $\uparrow$
& Yield/Wait $\uparrow$ \\
\midrule
\multirow{2}{*}{ACT}
& Independent Imit. & 33.3\% & 42.7\% & 34.4\% \\
& \textbf{SAI} & \textbf{56.7\%} & \textbf{74.7\%} & \textbf{68.9\%} \\
\midrule
\multirow{2}{*}{Diffusion}
& Independent Imit. & 23.5\% & 35.2\% & 30.7\% \\
& \textbf{SAI} & \textbf{52.9\%} & \textbf{62.2\%} & \textbf{76.9\%} \\
\bottomrule
\end{tabular}
\end{adjustbox}
\end{wraptable}
SAI operates at the data-curriculum level and is not tied to a particular action decoder. To test backbone compatibility, we instantiate the same curriculum with two imitation-learning backbones, Action Chunking Transformer (ACT) and Diffusion Policy, while keeping the sensorimotor interface and task protocol fixed. We evaluate this comparison on Painting Transport, a contact-rich task that requires synchronized grasping, lifting, transport, and placement. As shown in Table~\ref{tab:policy_backbone}, SAI improves over Independent Imitation for both backbones. With ACT, SAI improves task success from 33.3\% to 56.7\%, Phase Sync. from 42.7\% to 74.7\%, and Yield/Wait from 34.4\% to 68.9\%. With Diffusion Policy, SAI also improves all three metrics, including a large gain in Yield/Wait from 30.7\% to 76.9\%. These results support the interpretation that SAI is a data-curriculum contribution rather than an artifact of a particular action decoder.

\section{Conclusions and Limitations}
\label{sec:conclusion}

We introduced \textit{Sequential Asymmetric Imitation} (SAI), a single-teleoperator curriculum for learning decentralized dual-robot manipulation without synchronized dual-operator demonstrations. SAI trains Robot A from solo demonstrations, trains Robot B against the deployed Robot-A policy, and uses sparse interventions to correct coordination failures. Across deformable spreading, shared-workspace collection, and rigid-object transport, SAI improves task success, phase synchronization, and partner-contingent yielding over independent imitation. These results suggest that coordinated multi-robot behavior can emerge from structured imitation data, without reinforcement learning, explicit inter-robot communication, or dense synchronized demonstrations.
\vspace{-0.3em}

\textbf{Limitations:} SAI assumes that each robot can learn useful standalone primitives, that partner progress is partially observable from local sensory inputs, and that failures can be corrected through meaningful single-agent interventions. Tasks requiring simultaneous correction of both robots may not fit the current A-centric design. The main limitation is failure-mode coverage: our Stage-3 intervention set is only about 30\% of the baseline dataset size. Although this improves \textit{Yield/Wait}, recovery has not saturated. More targeted interventions around rare delays and near-failure contact states may improve robustness, while excessive correction may bias the policy toward unnecessary waiting.

\clearpage
\bibliography{ref}

\clearpage

\appendix

    \section{Implementation Details and Reproducibility}
    \label{appendix:implementation}

        This section summarizes the policy architecture, observation and action spaces, and training hyperparameters used in our real-robot experiments. Unless otherwise stated, all policies use an ACT-style action chunking backbone with temporal history encoding and a cascaded whole-body action head.
        
        \subsection{Network Architecture Details}
        
        \textbf{Visual Encoder.}
        We use a frozen DINOv3 (ViT-Base) backbone~\cite{simeoni2025dinov3} to extract spatial features from RGB observations, including the shared top camera and egocentric wrist cameras. The feature maps are projected with a $1\times 1$ convolution and flattened into image tokens before entering the transformer encoder. Two-dimensional sinusoidal positional embeddings are added to preserve spatial information.
        
        \textbf{History encoder.}
        To provide temporal context for long-horizon decisions, we maintain a history window of $H=30$ steps. For efficiency, we select 12 historical frames with logarithmic temporal sampling, using denser samples near the current timestep and sparser samples further in the past. For each sampled timestep, multi-view visual features are pooled and projected, while the full-body state is encoded by an MLP into a 64-dimensional latent. The visual and state latents are concatenated, passed through a fusion MLP, and encoded by a single-layer GRU with hidden dimension 512. Padding masks are used when historical frames are unavailable. The final GRU hidden state is used as the history token $z_{\mathrm{hist}}^t$.
        
        \textbf{Cascaded whole-body action head.}
        The ACT decoder outputs action-query embeddings. Instead of predicting all action dimensions with a single flat head, we use a cascaded structure. A 2-layer MLP first predicts the mobile-base action $a_{\mathrm{base}}\in\mathbb{R}^3$, corresponding to $(v_x,v_y,\omega)$. The predicted base intent is then concatenated with the decoder query and passed to a second 2-layer MLP to predict the arm-gripper action $a_{\mathrm{arm}}\in\mathbb{R}^{14}$. This factorization conditions arm and gripper commands on the imminent base motion and improves whole-body action consistency.

    \subsection{State and Action Space Definitions}

        \begin{wraptable}[18]{r}{0.45\linewidth}
    \centering
    \vspace{-0.45em}
    \captionsetup{font=normalsize,skip=2pt}
    \caption{Hyperparameters for policy instantiation.}
    \label{tab:hyperparameters}
    \footnotesize
    \setlength{\tabcolsep}{3pt}
    \renewcommand{\arraystretch}{0.88}
    \begin{adjustbox}{max width=\linewidth}
    \begin{tabular}{lc}
    \toprule
    Hyperparameter & Value \\
    \midrule
    \multicolumn{2}{c}{\textit{Architecture}} \\
    Visual backbone & Frozen DINOv3 \\
    History window & 30 steps \\
    Log. sampling steps & 12 steps \\
    State encoder dim. & 64 \\
    GRU hidden dim. & 512 \\
    ACT model dim. & 512 \\
    ACT enc./dec. layers & 4 / 1 \\
    Attention heads & 8 \\
    \midrule
    \multicolumn{2}{c}{\textit{Obs. and Action}} \\
    Proprio. state dim. & 17 $(3+14)$ \\
    Physical feedback & Arm torque \\
    Action horizon & 100 \\
    Action dim. & 17 $(3+14)$ \\
    \midrule
    \multicolumn{2}{c}{\textit{Training}} \\
    Temp. ensemble coeff. & 0.01 \\
    Optimizer & AdamW \\
    Learning rate & $1\times 10^{-5}$ \\
    Weight decay & $1\times 10^{-4}$ \\
    Visual backbone & Frozen \\
    \bottomrule
    \end{tabular}
    \end{adjustbox}
    \vspace{-0.35em}
\end{wraptable}

        \textbf{Observation space.}
        At each timestep, the observation consists of RGB images, robot proprioception, and torque feedback. RGB images are captured from the shared top-view camera and egocentric wrist cameras. The proprioceptive state is 17-dimensional, containing the 3D base state and 14D arm-gripper state. Torque feedback consists of measured arm-side torque signals, which provide implicit physical interaction cues during contact-rich manipulation.
        
        \textbf{Action space.}

        The policy predicts an action chunk of horizon $K$. Each action step has dimension $d_a=17$, consisting of a 3D base command $(v_x,v_y,\omega)$ and a 14D arm-gripper command. During execution, overlapping chunks are combined with exponential temporal ensembling using coefficient $\alpha=0.01$.
        
        \subsection{Training Details and Hyperparameters}
        
        All policies are trained with AdamW. During Stage 3, interventions may correct only part of the action space, such as stopping the base while maintaining arm posture. We therefore use binary action-loss masks $M^t\in{0,1}^{K\times d_a}$ for intervention samples, so gradients are applied only to explicitly corrected action dimensions. This prevents sparse corrective labels from overwriting nominal behavior in unaffected dimensions. Key hyperparameters are summarized in Table~\ref{tab:hyperparameters}.
\subsection{Data Budget and Stage-3 Interventions}
\label{app:intervention_statistics}

To control for training-data quantity, all methods use 80 trajectories per
robot. SAI allocates Robot A's data as 56 Stage-1 trajectories and 24
Stage-3 correction trajectories, while Robot B uses the same 80 Stage-2
trajectories as Serial Imitation.

\begin{table}[h]
\centering
\caption{Training trajectory allocation for Robot A / Robot B.}
\label{tab:data_budget}
\begin{tabular}{lcccc}
\toprule
Method & Stage 1 & Stage 2 & Stage 3 & Total \\
\midrule
Independent Imit. & 80 / 80 & 0 / 0 & 0 / 0 & 80 / 80 \\
Serial Imitation  & 80 / 0  & 0 / 80 & 0 / 0 & 80 / 80 \\
SAI               & 56 / 0  & 0 / 80 & 24 / 0 & 80 / 80 \\
\bottomrule
\end{tabular}
\end{table}

The Stage-3 data therefore changes the supervision distribution rather than
increasing the total trajectory budget. We further quantify how much active
human correction is required within these Stage-3 rollouts.

\begin{table}[h]
\centering
\caption{Stage-3 correction burden and dominant correction phases.}
\label{tab:intervention_statistics}
\begin{tabular}{lccc}
\toprule
Task & Corr. traj. & Active time & Dominant phase(s) \\
\midrule
Bed Throw  & 24 & $14.2\pm3.1$ s (12.5\%) & Transport (44.8\%) \\
Tablecloth & 24 & $8.1\pm2.5$ s (7.8\%) & Spread (71.7\%) \\
Laundry    & 24 & $7.4\pm1.8$ s (9.7\%) & Approach (58.3\%) \\
Painting   & 24 & $6.3\pm2.6$ s (9.1\%) &
Transport/lift (39.0/31.4\%) \\
\bottomrule
\end{tabular}
\end{table}

Active intervention occupies only 7.8--12.5\% of each correction rollout
and is concentrated around coordination-critical phases.

    \section{Partner Masking and Randomization}
        \label{app:partner_masking}
        
        \paragraph{Offline partner-region detection.}
        During Phase 1, Robot A interacts with a compliant human partner. Since the deployment partner is Robot B, directly training on raw Phase-1 images may cause the policy to exploit human-specific appearance cues, such as clothing, skin color, body texture, or human body shape. We therefore apply an offline partner-region masking pipeline to reduce this morphology-induced visual shortcut.
        
        For each raw observation $o_A^t \in \mathbb{R}^{H \times W \times 3}$, we use an off-the-shelf instance segmentation model to obtain a binary human-region mask $\hat{m}^t \in \{0,1\}^{H \times W}$. To cover boundary artifacts, thin uncovered regions, and nearby shadows, we dilate the mask with a circular structuring element:
        \begin{equation}
            m^t = \hat{m}^t \oplus \mathcal{K}_{\rho},
        \end{equation}
        where $\mathcal{K}_{\rho}$ is a circular dilation kernel and $\rho$ is the dilation radius. This preprocessing is applied only during offline data loading and does not introduce additional policy parameters.
        
\begin{figure}[t]
            \centering
            \begin{minipage}{0.32\textwidth}
                \centering
                \includegraphics[width=\linewidth]{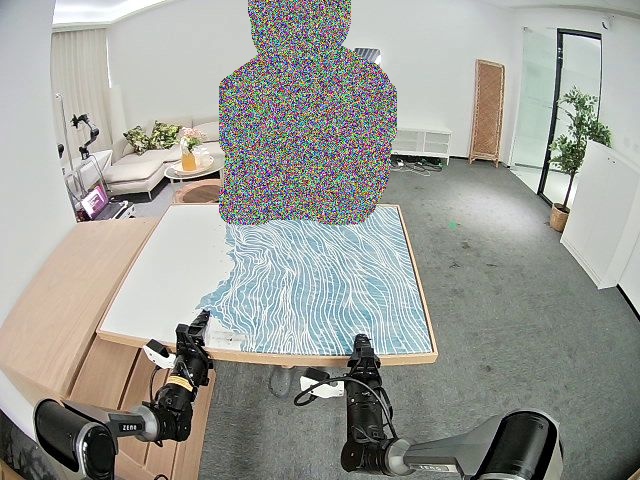}
                \vspace{-0.4em}
                \centerline{\small (a) RGB randomization}
            \end{minipage}
            \hfill
            \begin{minipage}{0.32\textwidth}
                \centering
                \includegraphics[width=\linewidth]{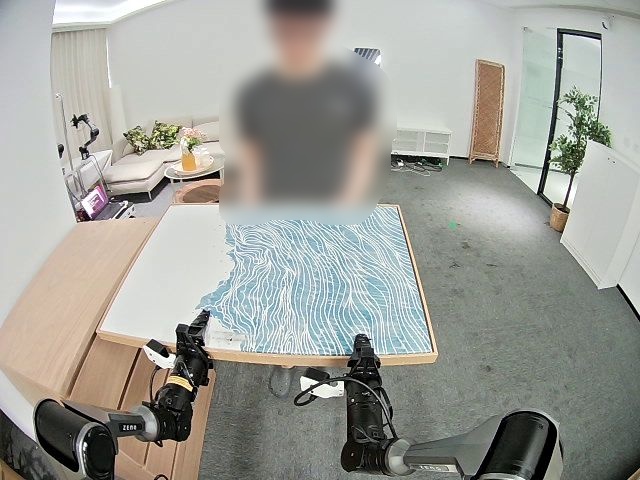}
                \vspace{-0.4em}
                \centerline{\small (b) Gaussian blur}
            \end{minipage}
            \hfill
            \begin{minipage}{0.32\textwidth}
                \centering
                \includegraphics[width=\linewidth]{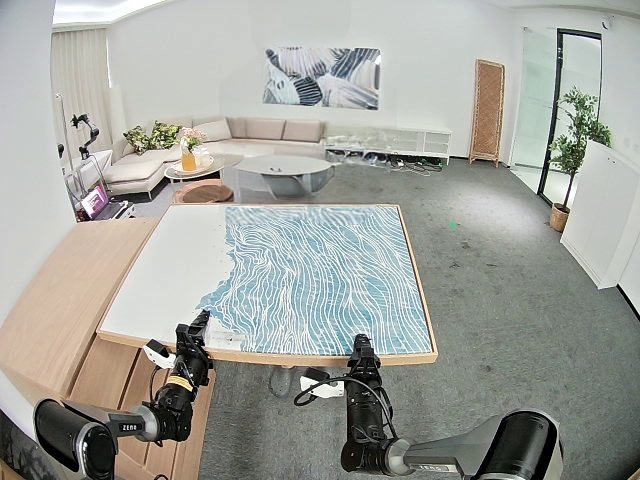}
                \vspace{-0.4em}
                \centerline{\small (c) Stable Diffusion inpainting}
            \end{minipage}
        
            \vspace{0.4em}
        
            \caption{
            \textbf{Partner-region randomization.}
            Training-time replacements applied inside the detected human-partner region: RGB randomization, Gaussian blur, and Stable Diffusion Inpainting.
            }
            \label{fig:desensitization_comparison}
        \end{figure}

        \paragraph{Partner-region randomization.}
        Given the dilated mask $m^t$, we construct an augmented observation:
        \begin{equation}
            \bar{o}_A^t
            =
            (1-m^t)\odot o_A^t
            +
            m^t\odot r(o_A^t),
        \end{equation}
        where $r \sim \mathcal{R}$ is a stochastic replacement operator applied only inside the masked partner region. The goal is not to remove all interaction evidence, but to make human-specific appearance cues unreliable during policy learning. This encourages the policy to rely less on the visual identity of the human partner and more on task-relevant scene cues, such as object geometry, deformation, contact progress, and workspace configuration. During offline training-time data loading, $\mathcal{R}$ randomly samples one of the following partner-region transformations:
        \begin{itemize}
            \item \textbf{RGB Randomization:} pixels inside the mask are replaced with independently sampled RGB values, disrupting local color and texture statistics.
            \item \textbf{Gaussian Blur:} a strong local blur is applied inside the mask to remove high-frequency appearance details while preserving only a coarse foreground region.
            \item \textbf{Stable Diffusion Inpainting:} the masked region is repaired using a mask-guided latent diffusion inpainting model~\cite{rombach2022high}. Unlike classical interpolation-based inpainting, this operator generates locally plausible content conditioned on the surrounding image context.
        \end{itemize}
        By varying the appearance of the partner region across these transformations, the policy is regularized against partner-appearance shortcuts. All transformations are applied offline during data loading and are not used during policy inference. Figure~\ref{fig:desensitization_comparison} visualizes the three partner-region transformations. These examples show how the detected human-partner region is altered during training while preserving the surrounding workspace context.
        
\section{Evaluation Protocol and Rollout Statistics}
\label{app:evaluation_protocol}

\paragraph{General protocol.}
All policies are evaluated on real-robot rollouts. The percentages reported in the main tables are computed from integer counts and rounded to one decimal place. \textit{Success} is a rollout-level metric. \textit{Phase Sync.} and \textit{Yield/Wait} are event-level descriptive metrics computed over predefined checkpoints or partner-contingent opportunities within the same rollout set. Therefore, their denominators can be larger than the number of rollouts, but the events should not be interpreted as independent rollout trials. Table~\ref{tab:rollout_statistics} summarizes the rollout count used for each main experiment.

\begin{wraptable}[7]{r}{0.58\linewidth}
\centering
\caption{Rollout statistics for the main experiments.}
\label{tab:rollout_statistics}
\scriptsize
\setlength{\tabcolsep}{4pt}
\renewcommand{\arraystretch}{0.95}
\begin{adjustbox}{max width=\linewidth}
\begin{tabular}{llcc}
\toprule
Experiment & Evaluation scope & Conditions & Rollouts / condition \\
\midrule
Q1 & Per-task comparison & 4 tasks $\times$ 3 methods & 30 \\
Q3 & Policy ablations & Failure-specific variants & 31 \\
Q4 & Backbone compatibility & Diffusion on Painting & 34 \\
\bottomrule
\end{tabular}
  \end{adjustbox}
\end{wraptable}

\paragraph{Q1: per-task evaluation.}
For the main per-task comparison, each task-method pair is evaluated over 30 rollouts. Success is computed at the rollout level with denominator 30. Phase Sync. is computed over task-specific synchronization checkpoints, and Yield/Wait is computed over task-specific partner-contingent opportunities. The number of checkpoints and opportunities is fixed per task and kept identical across methods. Table~\ref{tab:metric_denominators} lists the corresponding event-level denominators for each task.

\begin{table}[H]
\centering
\caption{Event-level denominators for Q1.}
\label{tab:metric_denominators}
\footnotesize
\setlength{\tabcolsep}{4pt}
\renewcommand{\arraystretch}{0.95}
\begin{tabular}{lcccc}
\toprule
Task
& Rollouts
& Phase ckpts.
& Phase denom.
& Yield denom. \\
\midrule
Bed-throw & 30 & 4 & 120 & 60 \\
Tablecloth & 30 & 3 & 90 & 60 \\
Laundry & 30 & 2 & 60 & 60 \\
Painting & 30 & 5 & 150 & 90 \\
\bottomrule
\end{tabular}
\end{table}

\paragraph{Phase Sync.}
For each task, we divide the cooperative process into ordered event stages. These stages define the checkpoints used to compute \textit{Phase Sync.}. The Phase Sync. checkpoints for each task are listed in Table~\ref{tab:task_stage_definitions}. For a task with $K$ checkpoints per rollout and $N$ evaluated rollouts, the denominator is $N \times K$. The numerator is the number of checkpoints for which both robots reach and complete the corresponding stage within the predefined timing tolerance and without causing a task-disrupting interaction. Thus, \textit{Phase Sync.} is computed as the percentage of synchronized checkpoints over all evaluated checkpoints. If a rollout fails before reaching a later stage, all unreached checkpoints are counted as failed. Stage definitions and timing thresholds are fixed per task and kept identical across all evaluated methods.
For example, Painting Transport has five checkpoints per rollout; with 30 rollouts, the Phase Sync. denominator is $30 \times 5 = 150$.

\paragraph{Yield/Wait.}
Yield/Wait measures whether a robot responds appropriately to a partner-contingent opportunity. A response is counted as correct when the robot slows down, waits, yields, adjusts pose, or resumes only after the partner becomes ready.
\paragraph{Manual scoring and scorer agreement.}
Phase Sync. and Yield/Wait are scored manually from the predefined
events above. Partner-response events are judged using a fixed
task-specific response window following a partner delay or phase mismatch.
To assess scorer dependence, a second scorer independently rescored a
randomized subset of the evaluation data. Agreement was 93.4\%, with
Cohen's $\kappa=0.856$.

\begin{table}[H]
\centering
\caption{Task-level event definitions for Phase Sync. evaluation.}
\label{tab:task_stage_definitions}
\scriptsize
\setlength{\tabcolsep}{4pt}
\renewcommand{\arraystretch}{1.05}
\begin{tabular}{lp{0.78\textwidth}}
\toprule
Task & Phase Sync. checkpoints \\
\midrule
Bed-throw Spreading
&
(1) Robot A hands the bed throw to Robot B, and Robot B establishes a stable grasp;
(2) both robots retreat from the handoff area while keeping the object controlled;
(3) the robots jointly move toward the bed region with matched timing and without excessive dragging;
(4) both robots place the bed throw at the target area in a coordinated manner. \\
\midrule
Tablecloth Spreading
&
(1) both robots establish stable grasps before spreading;
(2) the robots spread the tablecloth with compatible timing so that the cloth remains approximately aligned and tensioned;
(3) both robots complete final placement without premature release or large disturbance to the cloth. \\
\midrule
Laundry Collection
&
(1) the robots enter the shared workspace and approach the laundry item without blocking each other;
(2) the collection and delivery motion is completed while the robots maintain a compatible spatial order in the shared area. \\
\midrule
Painting Transport
&
(1) both robots establish stable grasps on the painting;
(2) they lift the painting together without excessive tilt;
(3) they enter the transport path while maintaining compatible relative positions;
(4) they transport the painting with stable coordination along the path;
(5) they lower and place the painting at the target location without premature release. \\
\bottomrule
\end{tabular}
\end{table}

\paragraph{Q3: policy-component ablations.}
The policy ablation study evaluates failure-specific rates over 31 targeted rollout segments. These segments are initialized near the task phases where the corresponding failures are most likely to occur, rather than from the beginning of the full task. Premature release is evaluated in the high-risk delivery/release phase of Laundry Collection and measures whether the robot releases the garment before reaching the valid receiving or drop-off state. Premature lowering is evaluated in the high-risk transition phase of spreading or transport tasks and measures whether the arm enters a lowering or release phase before the mobile base has completed the corresponding transport phase. The history-token ablation is evaluated on premature release, while the cascaded-head ablation is evaluated on premature lowering.
\paragraph{Q4: backbone compatibility.}
For backbone compatibility, we evaluate the same SAI data curriculum with different imitation-learning backbones while keeping the task protocol, observation space, and action space fixed. The comparison is conducted on Painting Transport, following the task protocol used in Q1 for both ACT and Diffusion Policy. Success is computed at the rollout level, while Phase Sync. and Yield/Wait use the same Painting Transport event definitions described above.

\subsection{Partner-Informed Control Baselines}
\label{app:partner_controls}

We additionally evaluate two partner-informed controls on Tablecloth and
Painting: Diffusion Co-Policy (DCP) and Centralized Joint Diffusion Policy
(CJDP). Both use explicit inter-robot information exchange and are trained
using the same Stage-2 data as Serial Imitation. We evaluate each method
under clean communication and under a 100-ms communication delay with
10\% packet loss. SAI uses no explicit inter-robot communication.
\begin{table}[h]
\centering
\caption{
Partner-informed controls on Tablecloth and Painting.
DCP and CJDP use explicit inter-robot information exchange, whereas
SAI does not. Communication impairment applies a 100-ms delay and
10\% packet loss. Brackets report 95\% confidence intervals.
}
\label{tab:partner_controls}

\small
\renewcommand{\arraystretch}{1.15}

\begin{tabular*}{0.95\textwidth}{
@{\extracolsep{\fill}}
lccc
@{}}
\toprule
Method & Success & Phase Sync. & Yield/Wait \\
\midrule

\multicolumn{4}{c}{\textit{Tablecloth}} \\
\addlinespace[2pt]

Serial Imitation
& 60.0 [42.3, 75.4]
& 67.8 [52.3, 81.2]
& 61.7 [48.3, 75.0] \\

DCP
& 63.3 [45.5, 78.1]
& 70.0 [55.6, 83.3]
& 66.7 [51.7, 80.0] \\

DCP + impairment
& 56.7 [39.2, 72.6]
& 62.2 [47.8, 75.6]
& 58.3 [43.3, 73.3] \\

CJDP
& 70.0 [52.1, 83.3]
& 73.3 [58.9, 86.7]
& 71.7 [56.7, 86.7] \\

CJDP + impairment
& 53.3 [36.1, 69.8]
& 57.8 [42.2, 73.3]
& 55.0 [40.0, 70.0] \\

\textbf{SAI}
& \textbf{66.7 [48.8, 80.8]}
& \textbf{68.9 [54.4, 82.2]}
& \textbf{70.0 [55.0, 85.0]} \\

\midrule
\multicolumn{4}{c}{\textit{Painting}} \\
\addlinespace[2pt]

Serial Imitation
& 46.7 [30.2, 63.9]
& 67.3 [54.0, 80.0]
& 48.9 [34.4, 63.3] \\

DCP
& 60.0 [42.3, 75.4]
& 74.0 [62.0, 85.3]
& 67.8 [54.4, 80.0] \\

DCP + impairment
& 46.7 [30.2, 63.9]
& 63.3 [50.0, 76.0]
& 52.2 [37.8, 65.6] \\

CJDP
& 66.7 [48.8, 80.8]
& 76.7 [65.3, 86.7]
& 71.1 [57.8, 83.3] \\

CJDP + impairment
& 43.3 [27.4, 60.8]
& 58.7 [46.0, 71.3]
& 47.8 [34.4, 61.1] \\

\textbf{SAI}
& \textbf{56.7 [39.2, 72.6]}
& \textbf{74.7 [62.7, 85.3]}
& \textbf{68.9 [55.6, 82.2]} \\

\bottomrule
\end{tabular*}
\end{table}

Under clean communication, DCP and CJDP are competitive with SAI on these
tasks. Under communication impairment, both partner-informed controls
decrease across the reported coordination metrics. These results position
SAI as a communication-free alternative rather than as a method intended
to outperform partner-informed control under perfect communication.

\end{document}